\documentclass[runningheads]{llncs}
\usepackage{eccv}
\usepackage{eccvabbrv}

\usepackage{microtype}
\usepackage{graphicx}
\usepackage{setspace}
\usepackage{wrapfig}
\usepackage{booktabs}

\usepackage[accsupp]{axessibility}

\usepackage{hyperref}
\usepackage{orcidlink}

\def\squarebox#1{\hbox to #1{\hfill\vbox to #1{\vfill}}}
\def\boxit#1{\vbox{\hrule\hbox{\vrule\kern6pt
			\vbox{\kern6pt#1\kern6pt}\kern6pt\vrule}\hrule}}

\newcommand*\samethanks[1][\value{footnote}]{\footnotemark[#1]}

\author{Yihang Chen\inst{1}\thanks{Equal contribution}
\and
Tsai Hor Chan\inst{1}\samethanks
\and
Guosheng Yin\inst{1}
\and
Yuming Jiang\inst{2}
\and
Lequan Yu \inst{1}\thanks{Corresponding Author}
}

\authorrunning{Y.~Chen et al.}
\institute{The University of Hong Kong, Hong Kong SAR, China\\ \email{\{yhchen, hchanth\}@connect.hku.hk} \email{\{gyin, lqyu\}@hku.hk}\and
Wake Forest University School of Medicine\quad\email{yumjiang@wakehealth.edu} 
}

\usepackage{graphicx}
\usepackage{amsmath}
\usepackage{amsfonts}
\usepackage{xcolor}
\usepackage{caption}
\usepackage{mathtools}
\usepackage{bm}
\usepackage{soul}
\usepackage{algorithm}
\usepackage{algpseudocode}

\newcommand{\para}[1]{\vspace{.05in}\noindent\textbf{#1}}

\begin{document}
\title{cDP-MIL: Robust Multiple Instance Learning via Cascaded Dirichlet Process}

\maketitle

\begin{abstract}
Multiple instance learning (MIL) has been extensively applied to whole slide histopathology image (WSI) analysis.
The existing aggregation strategy in MIL, which primarily relies on the first-order distance (e.g., mean difference) between instances, fails to accurately approximate the true feature distribution of each instance, leading to biased slide-level representations.
Moreover, the scarcity of WSI observations easily leads to model overfitting, resulting in unstable testing performance and limited generalizability. 
To tackle these challenges, we propose a new Bayesian nonparametric framework for multiple instance learning, which adopts a cascade of Dirichlet processes (cDP) to incorporate the instance-to-bag characteristic of the WSIs.
We perform feature aggregation based on the latent clusters formed by the Dirichlet process, which incorporates the covariances of the patch features and forms more representative clusters.
We then perform bag-level prediction with another Dirichlet process model on the bags, which imposes a natural regularization on learning to prevent overfitting and enhance generalizability.
Moreover, as a Bayesian nonparametric method, the cDP model can accurately generate posterior uncertainty, which allows for the detection of outlier samples and tumor localization.
Extensive experiments on five WSI benchmarks validate the superior performance of our method, as well as its generalizability and ability to estimate uncertainties. 
Codes are available at \url{https://github.com/HKU-MedAI/cDPMIL}.
\end{abstract}

\section{Introduction}

Whole slide histopathology images (WSIs) provide enriched information for modern diagnosis of diseases.
However, manually traversing through the WSIs for clinical diagnosis would be challenging for pathologists as such images are in giga-pixels (e.g., with a typical size of 80,000 $\times$ 80,000).
Recently, the multiple instance learning (MIL) paradigm has demonstrated success in modeling WSIs with modern deep learning techniques. 
Multiple instance learning (MIL) assumes that the data can be divided into sets of instances, referred to as bags, and aims to learn the label of each bag, which consists of multiple instances, with supervision solely provided by the bag label \cite{andrews2002support}.
MIL is often known as a weakly supervised learning problem because pixel-wise annotations are limited due to the enormous size of WSIs.

Existing MIL-based methods on WSI analysis mainly focus on designing effective algorithms to learn the inter-relations of patches, where attention-based \cite{ilse2018abmil, li2021dsmil, yang2022remix, shao2021transmil, chen2021multimodal} and graph-based \cite{zheng2022GTNMIL, hou2022h2MIL, chan2023HEAT, chen2021patchGCN, chen2023TBMF, chan2023SUMSHINE2} methods demonstrate significant success. 
Effective aggregation algorithms are then developed to pool the patch-level features to bag-level for predictions on slides.
However, several challenges remain:
First, the aggregation methods adopted by existing paradigms based on first-order distance (e.g., the mean difference of centroids) fail to provide accurate estimation of the bag-level feature distribution, which hinders the downstream task performance based on these pooled features.
Second, the bag-level prediction model would mostly suffer from overfitting due to the limited observations of bags, leading to poor testing performance and generalizability.
Third, most existing methods are based on a deterministic (i.e., frequentist) framework, which fails to capture the epistemic uncertainty (i.e., predictive confidence of the fitted MIL model).

\begin{wrapfigure}{r}{0.48\textwidth}
    \vspace{-10mm}
    \centering
    \includegraphics[width=0.47\textwidth]{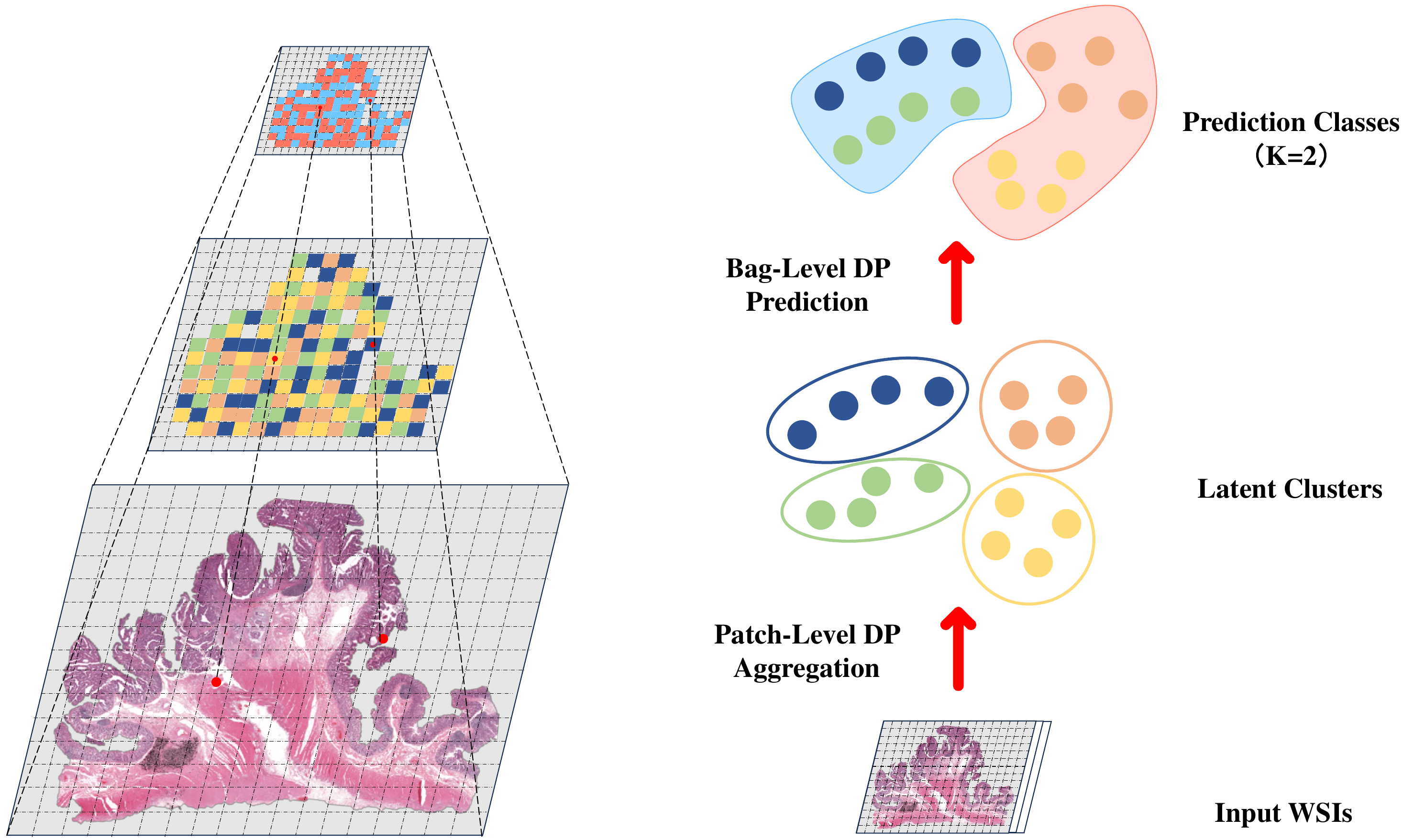}
    \caption{
    % Overview of Hierarchical DP on WSI.
    Illustration of a cascaded Dirichlet process model on whole slide images. 
    We first pool the WSI patches with latent Gaussian distributions, and then categorize the pooled WSIs into $K$ classes.
    }
    \label{fig: Framework_Intro}
    \vspace{-5mm}
\end{wrapfigure}
To tackle the limitations of the existing works, we propose a novel framework for MIL --- \underline{C}ascaded \underline{D}irichlet \underline{P}rocess \underline{M}ultiple \underline{I}nstance \underline{L}earning (cDP-MIL), which is a multi-stage Bayesian nonparametric framework learning the instance-level and bag-level representations. 
Fig. \ref{fig: Framework_Intro} illustrates our cDP design. 
We tackle instance-level feature pooling with a deep neural network parameterized Dirichlet process, which effectively clusters the local features.
The posterior log-likelihood estimated by the patch-level DP can also interpret the contribution of each patch to the slide-level prediction, which enables the localization of patches of interest (e.g., tumor patches).
We then design a DP predictive module to perform bag-level predictions, which can also quantify predictive uncertainties and generalize well to unseen data. 
Moreover, we parametrize the mean and covariance of each Gaussian component with a deep neural network and adopt a stochastic variational inference schema to optimize the parameters with gradient descent. 
This facilitates fast convergence and accurate approximation of the distributions in each component, leading to improved predictive performance. 
Experiments on five histopathology image datasets with cancer classification, subtyping, and localization tasks validate the superior performance of our proposed framework.
Thorough ablation analysis validates the contributions of the proposed modules and robustness to variations of key components and hyperparameters.
We also validate the capability of cDP-MIL to capture predictive uncertainty through out-of-distribution (OOD) detection and generalization tasks.

\section{Related Work}

\para{Bayesian Nonparametric Methods.}
Bayesian nonparametric methods aim to provide optimal solutions from an infinite dimensional parameter space.
They select the best subset of parameters over an infinite space to best explain finite sample observations.
There has been extensive development on Bayesian nonparametric methods in statistics, including Gaussian process (GP) \cite{rasmussen2006gaussian, li2020gp, haussmann2017VGPMIL, kandasamy2018neural,snoek2012practical}, Dirichlet process (DP) \cite{echraibi2020DPGMM, kandemir2014DPMIL, ishwaran2002exactDP, blei2006variationalDP}, and P\'{o}lya tree \cite{mauldin1992polya}.
Echraibi et al. \cite{echraibi2020DPGMM} combine DP with the Gaussian mixture model (GMM) and update the prior parameters and variational posterior in closed forms owing to the reparameterization trick \cite{kingma2013auto}. 
DP also shows outstanding performance in clustering due to  its flexible determination of the number of clusters and robust cluster allocation mechanism (from the richer-gets-richer characteristic of DP, i.e., a sample has a higher probability of being allocated to the clusters that have more observations) \cite{ronen2022deepdpm, blei2006variationalDP, echraibi2020DPGMM}. 
Existing works widely adopt Markov chain Monte Carlo (MCMC) methods, e.g., Gibbs sampling, to fit the DP model, which is not scalable to modern deep learning frameworks as their time complexity is high.
Recent advances in variational inference \cite{blei2006variationalDP} make deep Bayesian nonparametric methods possible by introducing a surrogate variational objective from the probabilistic assumptions. 

\para{Multiple Instance Learning. }
Multiple Instance Learning (MIL) serves as a general framework for handling a series of objects, such as text instances, natural images, and medical images (e.g., WSIs) \cite{liu2023dsca,hou2022h2MIL,zheng2022GTNMIL,chen2021multimodal,chen2021patchGCN,graham2019hover,wang2019rmdl}. 
Existing MIL paradigms divide WSIs into patches and perform slide-level prediction through pooling on instance-level features.
Attention-based MIL (ABMIL) \cite{ilse2018abmil} employs the attention mechanism to aggregate these features, enabling accurate predictions as well as providing interpretability.
In addition, there are approaches that model slides using graphs and obtain bag-level predictions through the utilization of graph convolutional methods \cite{chan2023HEAT, chen2021patchGCN, hou2022h2MIL, zhao2023mulgt, zheng2022GTNMIL}.
However, these methods are mostly deterministic and lack the ability to estimate the inherent uncertainty within the fitted model.
Furthermore, the design of regularization poses significant challenges, as these methods are often prone to overfitting.

\para{Feature Aggregation in MIL.}
Due to the weak supervision nature of MIL, it is difficult to accurately estimate the bag-level distribution from many unlabelled instances, thus making the design of an effective aggregation mechanism necessary.
Many existing works attempt to design a pooling module that effectively aggregates instance-level features into the bag level \cite{hou2022h2MIL, ilse2018abmil, yang2022remix, chan2023HEAT}.
HEAT \cite{chan2023HEAT} designs a pooling method based on the predicted patch type to mitigate imbalanced observations of normal and tumor patches. 
However, most of these methods operate on the first-order difference of the instances in a bag, such as the mean difference, and neglect the covariance structures of the feature embeddings. 
The predicted clusters may suffer from inherited observational bias and decrease the quality of aggregated representations.

\para{Bayesian Methods in MIL. }
Recently, Bayesian methods \cite{kandemir2014DPMIL, yufei2022bayes, haussmann2017VGPMIL} have emerged as a promising solution to addressing the aforementioned challenges. 
These methods offer natural regularization in the learning process while also providing the ability to quantify uncertainty by sampling from the posterior predictive distribution.
BayesMIL \cite{yufei2022bayes} designs a Bayesian attention framework for WSI classification by placing a mixture of log-normal distribution on instances to allow for uncertainty estimation and better interpretability.
However, most probabilistic assumptions (e.g., Gaussian distribution) in conventional Bayesian methods are strong and do not fit satisfactorily to large WSI analysis, prompting the need to design deep nonparametric methods.
Nonparametric Bayesian methods also excel in MIL due to their capability of feature representation learning \cite{blei2006variationalDP, ZhangChenyang2023Bnao,echraibi2020DPGMM, kandemir2014DPMIL, haussmann2017VGPMIL, kim2010gaussian, chan2024adaptive}. 
Some methods incorporate GP or DP into MIL learning, which, however, mainly focus on text analysis, so that their adaptivity to large-scale WSI analysis is unknown.
Furthermore, they are also limited to MCMC methods or traditional variational inference methods for posterior computation, which makes the computational complexity very high.
Moreover, these methods do not evaluate predictive uncertainty inherited from the models.

\section{Methodology}
\begin{figure}[t]
    \centering
    \includegraphics[width=0.96\textwidth]{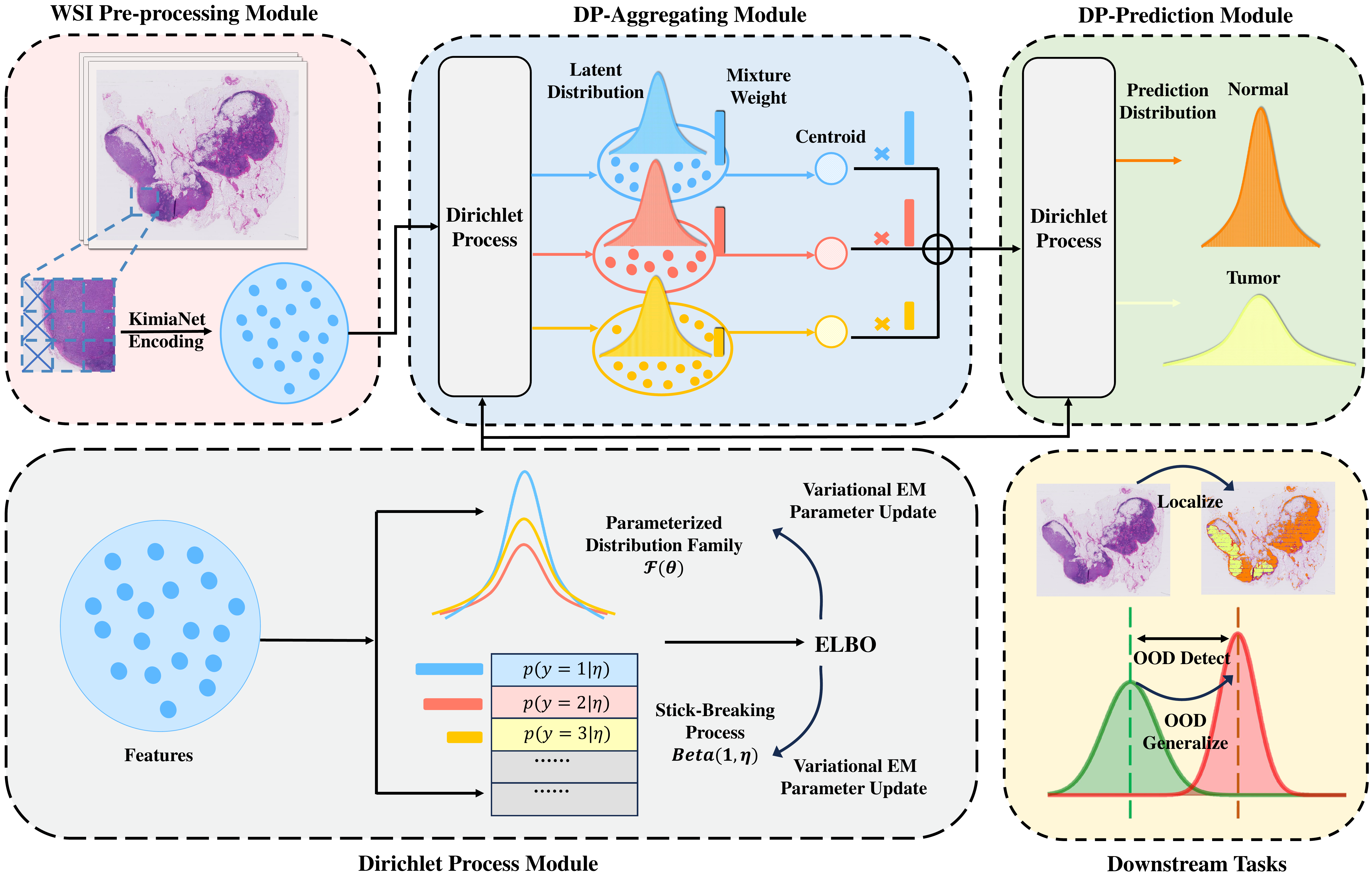}
    \caption{The overall framework of our proposed cascaded DP model. 
    We first use Otsu's segmentation to outline tissue regions and select foreground instances for MIL learning.
    We then design an aggregation module based on the Dirichlet process to cluster instance-level features into latent clusters.
    Finally, a prediction module based on DP would perform predictions based on the distribution learned for each cluster.
    }
    \label{fig: Framework_Method}
    \vspace{-3mm}
\end{figure}

In order to capture features of WSIs from different levels more effectively, our method consists of two major designs: an aggregation module based on DP to form latent clusters, and a DP classifier to extract the most representative features from latent clusters for bag-level predictions.
The combination of these two modules formulates a cascaded DP design which allows for more effective allocations of weights to features.
Fig. \ref{fig: Framework_Method} shows an overview of our proposed framework and Algorithm \ref{alg: cDP-MIL} presents the detailed workflow. 

\begin{algorithm}[]
\begin{algorithmic}[1]

\Statex \textbf{Input:} 
\Statex Number of bags $B$, number of instances per bag $\{N_1,\ldots,N_B\}$
\Statex Extracted features $\{\mathbf{X}_b\}_{b=1}^B$, bag-level labels ${\mathbf{ Y}}=\{Y_1,\ldots,Y_B\}$,
\Statex Hyperparameters $\{\mathbf{V}_0, \mathbf{m}_0, \kappa, T, K, \eta_1, \eta_2, p\}$, 
\Statex Parameter set of network $\mathbf{\Psi} = \{ \psi_t\}_{t=1}^T$ and $\mathbf{\Upsilon} = \{ \mathbf{\upsilon}_k\}_{k=1}^K$ to encode mean and covariance.
\Statex \textbf{Output:} 
\Statex Trained MIL model $\mathcal{M}$

\State Initialize $\varphi \in \mathbb{R}^{B\times K}$
\State Initialize $\mathbf{\Phi}=[\mathbf{\phi}_1,\ldots,\mathbf{\phi}_B],$ with $ \mathbf{\phi}_b \in \mathbb{R}^{N_b \times T}$
\For{each training epoch}
\For{each bag $b$}
\State $\mathbf{\phi}_b \leftarrow $ Fit-DP (${\mathbf{ X}}_b, \mathbf{\phi}_b, \mathbf{\Psi},T, \eta_1$) \Comment{Pooling by Algorithm \ref{alg: dp-pooling}}
\State Cluster assignments $\mathbf{Z}_b =\{z_{bj}=\mathop{\arg\max}\limits_{t} \phi_{bjt}, \  j\leq N_b, t\leq T\}$
\State Centroids $\mathbf{C}_b=\text{concat}(\text{mean}\{\mathbf{x}_{bj}, z_{bj}=t\}, t\leq T ) \in \mathbb{R}^{T \times p}$ 
\EndFor
\State $\mathbf{C} \leftarrow  \text{Concatenate}(\{\mathbf{C} _b\})$
\State $\varphi \leftarrow $ Fit-DP $(\mathbf{C}, \varphi,  \mathbf{\Upsilon}, K ,\eta_2, \mathbf{Y})$ \Comment{Prediction by Algorithm \ref{alg: dp-pooling}}
\EndFor

\State \Return trained $\mathcal{M}$.

\end{algorithmic}
\caption{Our proposed cDP-MIL framework}
\label{alg: cDP-MIL}
\end{algorithm}
\subsection{Preliminaries}

\para{Multiple Instance Learning (MIL).}
Given the $b$-th bag of instances (e.g., WSIs) $\mathbf{X}_b = (\mathbf{x}_{b1}, \ldots, \mathbf{x}_{bN_b})$, where $\mathbf{x}_{bk} \in \mathbb{R}^d$ is the $d$-dimensional embedding of the $k$-th instance, $N_b$ is the number of instances of the $b$-th bag, and the $b$-th bag has label $Y_b$.
The dataset $\mathcal{D} = \{\mathbf{X}_b, Y_b\}_{b=1}^B$ is composed of $B$ such bag-level pairs.
The objective of MIL is to learn an optimal model $\mathcal{M}$ for predicting the bag-level label with the bag of instances as input.

\subsection{Cascaded Dirichlet Process for WSI Analysis}
\begin{wrapfigure}{r}{0.52\textwidth}
\vspace{-10mm}
    \centering
\includegraphics[width=0.5\textwidth]{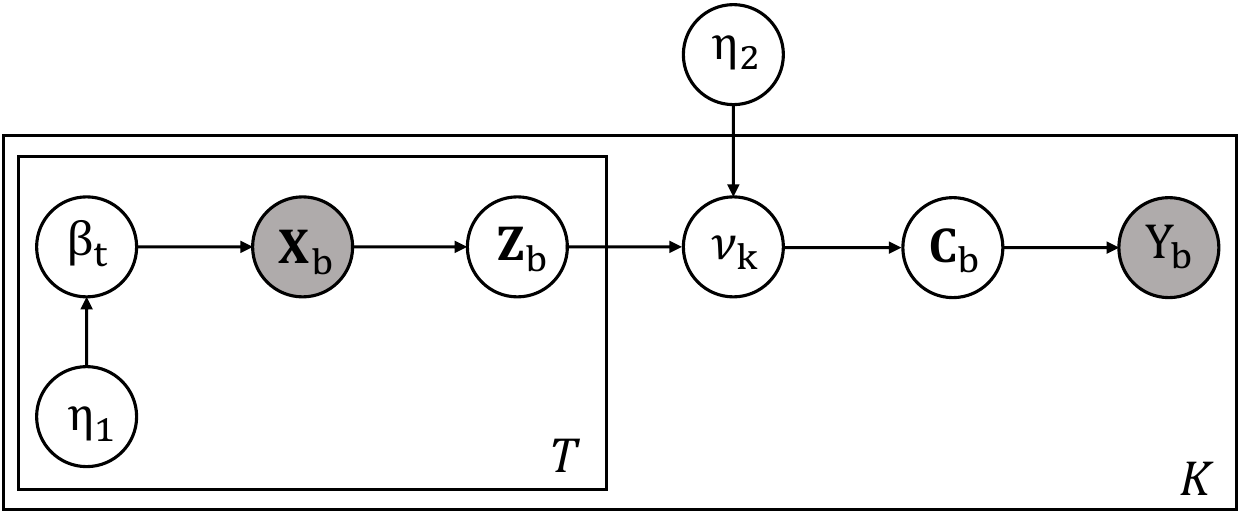}
    \caption{The probabilistic model of the cascaded DP in plate notation on the $b$-th bag.}
    \label{fig: plate-notation}
    \vspace{-20mm}
\end{wrapfigure}

We design a cascaded Dirichlet process model for learning the instance-level and bag-level representations. Instance-level aggregation is performed using the first DP module by assuming that similar patches have the same latent cluster label.  We then aggregate cluster features into bag-level features, where new latent cluster labels correspond to our task directly (e.g., each cluster represents tumor/normal for binary cancer classification).
Fig.~\ref{fig: plate-notation} presents our proposed model in plate notation. 

\para{Dirichlet Process. }
A Dirichlet process, denoted as $\text{DP}(\alpha, H)$, is a random probability measure on the sample space $\mathcal{X}$, such that for any measurable finite partition $S=\{B_i\}_{i=1}^K$ of the sample space,
\begin{align*}
    (X(B_1), X(B_2), \ldots, X(B_K)) \sim \text{Dir}(\alpha H(B_1), \alpha H(B_2) \ldots, \alpha H(B_K))
\end{align*}
where $\alpha$ is a positive concentration rate, $H$ is the base probability measure and $\text{Dir}(\cdot)$ stands for Dirichlet distribution. 

\para{DP Aggregation Module on Patch Level.}
With the features extracted for each instance, we design an aggregation mechanism based on DP to aggregate instance-level features to the bag level.
We approximate DP using the stick-breaking formulation, i.e., for the $b$-th WSI,
\begin{align} \label{eq: stick-breaking}
    \beta_{t} \sim \text{Beta}(1, \eta_1), \quad
     \pi_{t} = \beta_t \prod_{l=1}^{t-1} (1 - \beta_l), 
\end{align}
where $\pi_{t}$ is the probability assigned to each cluster of similar patches.
We have $z| \mathbf{\pi} \sim \text{Cat}(\cdot | \mathbf{\pi})$, characterizing the cluster assignment from a multinomial distribution based on the cluster probability $\mathbf{\pi}$.
Let $\mathbf{\theta}_{t} = (\mathbf{\mu}_{t}, \mathbf{\Sigma}_{t})$ be the mean and variance parameters of the normal density of the $t$-th cluster, and we place a normal-inverse-Wishart (NIW) prior on $\mathbf{\theta}_t$. 
We assume a Wishart prior on the precision matrix $\mathbf{\Lambda}_t$ (i.e., $\mathbf{\Lambda}_t = {\mathbf{\Sigma}_t}^{-1}$),
\begin{align*}
    f_{\mathbf{\Lambda}_t} (\mathbf{\Lambda}_t; \kappa, {\mathbf{V}_t}) = \dfrac{1}{2^{\kappa p/2} |\mathbf{V}_t|^{\kappa/2} \Gamma_p \left({\kappa}/{2}\right)}| \mathbf{\Lambda}_t|^{(\kappa-p-1) / 2} e^{-\text{tr}({\mathbf{V}_t}^{-1} \mathbf{\Lambda}_t)/2},
\end{align*}
where $\kappa > 0$, $\mathbf{\Lambda}_t$ is a positive semi-definite matrix and $\mathbf{V}_t \in \mathbb{R}^{p \times p}$ is a fixed symmetric positive definite matrix, and $\Gamma_p$ is the multivariate gamma function.
Given the covariance matrix,  $\mathbf{\theta}_t$ is distributed as NIW, with the joint density function 
\begin{align} \label{eq: NIW}
    p(\mathbf{\mu}_t, \mathbf{\Sigma}_t; \kappa, \mathbf{m}_t,  \mathbf{V}_t) = \mathcal{N}\left(\mathbf{\mu}_t; \mathbf{m}_t, \mathbf{\Sigma}_t/\kappa \right) f_{\mathbf{\Lambda}_t} (\mathbf{\Lambda}_t; \kappa, \mathbf{V}_t).
\end{align}
Because it is computationally infeasible to model an infinite number of clusters posited by the DP, we truncate the number of DP clusters in each bag by $T$.
The statistical properties of the mixed distribution are not affected by the truncation 
%according to Blei \etal~ 
\cite{blei2006variationalDP} as corroborated by Theorem 1 in the supplementary materials. 

\para{DP Prediction Module on Slide Level.}
Patches in a bag are fitted to the respective latent clusters of that slide after the DP aggregation module.
To estimate predictive uncertainty, we design a DP predictive module to classify the centroids of the clusters into $K$ classes (the maximum number of clusters is truncated at $K$).
We define a stick-breaking process for the bag-level DP as 
\begin{align} \label{eq: stick-breaking}
    \nu_k \sim \text{Beta}(1, \eta_2), \quad
     \pi_k = \nu_k \prod_{l=1}^{k-1} (1 - \nu_l).
\end{align}
The bag-level DP learns $K$ mixtures of Gaussian distributions, where each mixture component has an NIW prior distribution as defined in Eq. (\ref{eq: NIW}).
The classification can be performed by predicting the cluster assignment of each centroid with the largest log responsibility (i.e.,  the log-likelihood weighted by the DP mixture weights, see Eq. (1) in the supplementary materials) and then taking the average on different centroids.

\subsection{Posterior Computation via Variational Inference}
Directly inferring the cDP model through MCMC (e.g., Gibbs sampling) would be infeasible since the time complexity is high. 
Therefore, we adopt a variational inference procedure to compute the posterior distribution of the cDP.

\begin{algorithm}[]
\begin{algorithmic}[1]

\Statex \textbf{Input:} 
\Statex Features $\mathbf{X} \in \mathbb{R}^{n \times p}$, prior $\mathbf{\phi}$, encoder network $\mathbf{\Psi}$,
\Statex  Number of components $T$, concentration parameter $\eta$, optional label $Y$
\Statex \textbf{Output: } Updated log-responsibility $\mathbf{\phi}$
\State Initialize $\mathbf{\gamma}_1, \mathbf{ \gamma}_2$
\Repeat
\State Encode mean $\mu_{\mathbf{\Psi}}$ and covariance $\mathbf{\Sigma}_{\mathbf{\Psi}}$ with $\mathbf{\Psi}$
\State Update $\gamma_{1,t} \leftarrow  1 + \sum_{j=1}^{n} \phi_{jt}, \quad \gamma_{2,t} \leftarrow \eta + \sum_{j=1}^{n} \sum_{r=1}^{t} \phi_{jr} $
\State $\mathbb{E}_{\beta \sim q}[\log \pi_t] \leftarrow \text{digamma}(\gamma_{1, t}) -\text{digamma}(\gamma_{1, t} + \gamma_{2, t}) $
\State $\phi_{jt} \leftarrow \mathbb{E}_{\beta \sim q}[\log \pi_t] + \mathbb{E}[\log p(\mathbf{x}_j)] + \mathbb{H}[q_{\psi_t}(\cdot| z_j=t, \mathbf{x}_j)]$
\State Compute ELBO by Eq. (\ref{eq: elbo})
\State Backpropagate ELBO to $\mathbf{\Psi}$
\Until{convergence}
\State \Return $\mathbf{\phi}$
\end{algorithmic}
\caption{Fit Dirichlet Process}
\label{alg: dp-pooling}
\end{algorithm}

\para{Parameterization. }
Unlike the conventional deep DP methods that heavily rely on the parameterization tricks to learn a diagonal covariance matrix \cite{echraibi2020DPGMM, kandemir2014DPMIL}, we learn the full covariance matrix for better approximation of the distribution of each component. 
We parameterize the mean and covariance matrix with deep neural networks --- a network $\psi_t$ that encodes both the mean $\mu_{\psi_t}(x_i)$ and covariance matrix ${\mathbf{\Sigma}}_{\psi_t}(x_i)$ for each component $t$.
We then obtain $\mathbf{\Psi} = \{\psi_t\}_{t=1}^T$ as the set of trainable parameters.
This enables the bag label information to be backpropagated to the mean and covariance of the DP for clustering through variational inference.
Furthermore, the neural network generates distinct mean and covariance for each input sample. 
This allows for inductive distribution learning that the model can generate proper means and covariance matrices for unseen samples.

\para{Variational Objective. }
Since the posterior of the proposed DP is intractable (i.e., no closed-form solutions), we adopt a variational posterior $q$ to approximate the true posterior distribution with the variational posterior.
We proceed with a mean-field variational inference by factorizing the joint distribution into marginal distributions.
The variational family of the DP model that we optimize during variational inference is given by
\begin{align*}
    q(\mathbf{\beta}, \mathbf{\nu}, \mathbf{\theta}, \mathbf{\Lambda}, z) =  \prod_{t=1}^{T-1}q(\beta_t) \prod_{k=1}^K q(\nu_k) \prod_{t=1}^T q(\mathbf{\theta}_t) \prod_{t=1}^{T}q(\mathbf{\Lambda}_t) \prod_{b=1}^B q(z_b).
\end{align*}
The surrogate loss of backpropagation can be obtained by negating the evidence lower bound (ELBO), which is computed by the log responsibility 
and the KL divergence of the Wishart prior,
\begin{equation} \label{eq: elbo}
\begin{split}
    \mathcal{L}(\mathbf{\Psi}) =& \text{KL}(q(\mathbf{\beta}) \| p (\mathbf{\beta})) + \text{KL} (q(\nu) \| p(\nu))+\text{KL}(q(\mathbf{\theta}) \| p (\mathbf{\theta}))  \\& +\text{KL}(q(\mathbf{\Lambda}) \| p (\mathbf{\Lambda})) + \sum_b \text{KL}(q(z_b) \| p (z_b)),
\end{split}
\end{equation}
where $\text{KL}(q(z_b) \| p (z_b))$ can be interpreted as the supervised loss (e.g., cross-entropy loss for classification). 
If no supervision label is provided for the $b$-th sample, $\text{KL}(q(z_b) \| p (z_b))$ can be set as zero. 
Then the problem is reduced to learning the likelihood of inputs to the assumed DP model.
The detailed parameter update procedure of the variational posterior and the closed-form KL divergences of the assumed distributions can be found in the supplementary materials. 

\para{Prediction. }
We derive the predictive distribution of the Dirichlet process model \cite{echraibi2020DPGMM} at patch-level as 
\begin{equation} \label{eq: posterior predictive}
\begin{split}
    p(z = t  | \mathbf{x})&\propto  \int p(\mathbf{x}| \mathbf{\theta}_t) q(\mathbf{\theta}_t |z=t) d\mathbf{\theta}_t\\
     & \propto \mathbb{E}_{\beta \sim q(\beta)} [\pi_t(\beta)] \times \mathbb{E}_{q_{\psi_t}}[p(\mathbf{x}| z=t)],
\end{split}
\end{equation}
where $\mathbf{x}$ is the feature embedding of the testing sample, $p(\cdot)$ refers to Gaussian density function and $z$ refers to the cluster assignment. 
We obtain the logits from this posterior distribution and combine them with the logits from the labeled data.
We can design a DP predictive module from Eq. (\ref{eq: posterior predictive}) and use the ELBO to train $\mathbf{\Psi}$ for patch-level prediction. 
Similarly, slide-level prediction can be performed by Eq. (\ref{eq: posterior predictive}) with respective parameters of the slide-level DP.

\subsection{Patch Score and Uncertainty Estimation}
Patch scores and model uncertainty can be defined using patch-level DP and slide-level DP, respectively. For tumor/normal binary classification, the log-likelihood output by patch-level DP could be used as the score for each patch,
\begin{align} \label{eq: patch-scores1}
p(\mathbf{x}_j) = \sum_{t=1}^T \phi_t \mathbb{E}_{q_{\psi_t}}[p(\mathbf{x}_j| z_j=t)].
\end{align}
indicating how likely the patch is connected with tumors, where $\mathbf{x}_j$ is the feature embedding and $z_j$ is the latent cluster assignment of the $j$-th patch. Patch-level scores have great interpretability because they can help to localize the tumor region.
Adopting a deep Bayesian nonparametric framework allows for uncertainty estimation through posterior predictive distributions.
We adopt the log-likelihood from DP as the measure of uncertainty.
The log-likelihood of the slide-level DP can be defined by the posterior mixture density function of the DP, 
$
    p(\mathbf{x}_b) = \sum_{k=1}^K \phi_k \mathbb{E}_{q_{\psi_k}}[p_X(\mathbf{X}_b| z_b=k)],
$
where $\mathbf{X}_b$ is the feature embedding and $z_b$ is the latent cluster assignment of the $b$-th slide.
\begin{wraptable}{r}{0.55\textwidth}
    \vspace{-11mm}
    \centering
    \caption{Summary of datasets under two different tasks.}
    % \resizebox{\linewidth}{!}{ \def\arraystretch{1.05}
    \begin{tabular}{lcccc}
    \toprule
    \multicolumn{1}{l}{\textit{Classification}} &
    \multicolumn{2}{c}{Tumor} &
    \multicolumn{2}{c}{Normal} \\ \hline
    \multicolumn{1}{l}{\textbf{TCGA--COAD}} & \multicolumn{2}{c}{863} & \multicolumn{2}{c}{109} \\% & 1424\\
    \multicolumn{1}{l}{\textbf{TCGA--BRCA}} & \multicolumn{2}{c}{1569}  & \multicolumn{2}{c}{394} \\% & 1712\\ 
    \multicolumn{1}{l}{\textbf{Camelyon 16}} & \multicolumn{2}{c}{160} & \multicolumn{2}{c}{239}\\ \hline
    \multicolumn{1}{l}{\textit{Subtyping}} &
    \multicolumn{2}{c}{Type I} &
    \multicolumn{2}{c}{Type II} \\ \hline
    \multicolumn{1}{l}{\textbf{TCGA--ESCA}} & \multicolumn{2}{c}{131}  & \multicolumn{2}{c}{107} \\
    \multicolumn{1}{l}{\textbf{TCGA--BRCA}} & \multicolumn{2}{c}{283}  & \multicolumn{2}{c}{1420} \\
    \multicolumn{1}{l}{\textbf{TCGA--NSCLC}} & \multicolumn{2}{c}{540}  & \multicolumn{2}{c}{512} \\
    \bottomrule
    \end{tabular}
    \vspace{-22mm}
    \label{tab: Datasets summary}
\end{wraptable}

\section{Experiments}
\subsection{Experiments Setup}
\para{Datasets and Tasks.} 
We evaluate the performance of our method on five WSI analysis benchmarks --- TCGA--COAD, TCGA--BRCA, TCGA--ESCA, and TCGA--NSCLC from the TCGA project \cite{weinstein2013TCGA} and Camelyon 16 \cite{bejnordi2017camelyon16}. 
On average, each WSI in the TCGA datasets consists of 2231 patches at magnification level 20 (24401 patches for each WSI in Camelyon 16). 
We evaluate our method on cancer subtyping and cancer classification tasks. 
For the cancer subtyping task, all WSIs in the TCGA--ESCA dataset are assigned the label ``Type I: adenocarcinoma'' or the label ``Type II: squamous cell carcinoma'' while the WSIs in TCGA-BRCA are assigned ``Type I: Ductal Carcinoma'' or ``Type II: Lobular Carcinoma''. As for TCGA--NSCLC, there are two subtypes of cancer, namely, ``Type I: Lung Adenocarcinoma (LUAD)'' and ``Type II: Lung Squamous Cell Carcinoma (LUSC)''. 
For the cancer classification task, all the cases are divided into ``Normal'' and ``Tumor'' classes. We use 10-fold cross-validation to validate our model on these datasets. Table \ref{tab: Datasets summary} presents a summary of the datasets.
Additional training details and data split can be found in the supplementary materials.

In general, models potentially capture the epistemic and aleatoric uncertainties well, if they can satisfactorily detect the OOD samples and generalize performances on these samples. We validate the capability of uncertainty estimation of our method with OOD detection tasks, and the generalizability of our method on OOD generalization tasks.
For the OOD detection task, we compare with common measures of uncertainty such as maximum confidence, entropy, and differential entropy.
Specifically, we treat COAD as the in-distribution dataset and BRCA as the OOD dataset, and adopt the area under the receiver operating curve (AUROC) and the area under the precision-recall curve (AUPR) as the evaluation metrics. 
For OOD generalization, we train models on the COAD dataset and validate on the BRCA dataset on the cancer binary classification task.

\para{Preprocessing and Petraining.}
%-----------------------------------
We adopt the commonly used Otsu's thresholding algorithm \cite{chen2021patchGCN} to outline the tissue regions and use sliding windows to crop each WSI into non-trivial patches (trivial patches are mostly background).
We then use a pretrained encoder to extract the features of each patch.
Specifically, we adopt a SimCLR pretrained KimiaNet \cite{zheng2022GTNMIL} on 7,126 whole slide image datasets to generate the patch features $\{\mathbf{X}_b\}_{b=1}^B$. 

\para{Evaluation Metrics. } We use accuracy, F1-score, and AUROC to evaluate the classification performance.
Percentage values are used for the reported metrics with standard deviations in brackets. For all metrics, a higher value in percentage or a lower value in standard deviation indicates a more accurate or stable method.
\begin{wraptable}{r}{0.51\textwidth}
    \vspace{-10mm}
    \centering
    \caption{ LUSC/LUAD subtypes classification results [\%] of our cDP-MIL and other methods on the TCGA-NSCLC dataset.}
    \begin{tabular}{lcc p{1.3cm}}
    \toprule
    % \multicolumn{1}{l}{\textbf{}} &\\
    \multicolumn{1}{l}{\textbf{Model}} &
    \multicolumn{1}{c}{\textbf{Accuracy}} &
    \multicolumn{1}{c}{\textbf{AUROC}}  \\
    \hline
    ABMIL \cite{ilse2018abmil} & 77.2 & 86.6 \\
    DSMIL \cite{li2021dsmil} & 80.6 & 89.3 \\
    CLAM-SB \cite{lu2021CLAM} & 81.8 & 88.2 \\
    CLAM-MB \cite{lu2021CLAM} & 84.2 & 93.8 \\
    TransMIL \cite{shao2021transmil} & 88.4 & 96.0 \\
    \hline
    \textbf{cDP-MIL} & \textbf{93.5} & \textbf{97.4}\\
    \bottomrule
    \end{tabular}
    \vspace{-8mm}
    \label{tab: typing_NSCLC}
\end{wraptable}

\para{Competitive Methods.} We make comparisons with a variety of competitors, including ABMIL \cite{ilse2018abmil}, DSMIL \cite{li2021dsmil}, PatchGCN \cite{chen2021patchGCN}, GTNMIL \cite{zheng2022GTNMIL}, H$^2$-MIL \cite{hou2022h2MIL}, BayesMIL \cite{yufei2022bayes}, Remix \cite{yang2022remix}, CLAM \cite{lu2021CLAM}, TransMIL \cite{shao2021transmil}, and HEAT \cite{chan2023HEAT},  with detailed descriptions of each method given in the supplementary materials.

\subsection{Quantitative Results on Predictive Performance}
We first evaluate cDP-MIL by slide-level classification. 
Tables \ref{tab: typing_NSCLC}, \ref{tab: results_typing_ESCA} and \ref{tab: cancer-classification} present the performance of our model on cancer subtyping and cancer classification.
Compared with other state-of-the-art (SOTA) WSI or MIL frameworks \cite{ilse2018abmil, li2021dsmil, zheng2022GTNMIL, hou2022h2MIL,  chan2023HEAT}, our proposed method shows outstanding performance on both tasks, indicating the capability of cDP-MIL to regularize the learning process and reduce the train-to-test generalization error.

Furthermore, comparing cDP-MIL with another Bayesian MIL framework, Bayes-MIL \cite{yufei2022bayes}, our method shows large improvements in these tasks, which is evidence of the possible strength that the model excels in leveraging inter-cluster information. Hence, these experiments demonstrate the superior ability of our method in capturing significant features (i.e., the richer-gets-richer property) of WSI and infer the possible latent distributions of features.

\begin{table}[h]
    \centering
         \caption{Cancer subtyping results [\%] of our method and other methods on the TCGA--ESCA and TCGA--BRCA datasets.}
         \scalebox{0.9}{
        \begin{tabular}{lccc|ccc p{1.3cm}}
        \toprule
        \multicolumn{1}{l}{} & \multicolumn{3}{c|}{\textbf{TCGA--ESCA}} & \multicolumn{3}{c}{\textbf{TCGA--BRCA}}\\
        \multicolumn{1}{l}{\textbf{Model}} &
        \multicolumn{1}{c}{\textbf{AUROC}} & 
        \multicolumn{1}{c}{\textbf{Accuracy}} &
        \multicolumn{1}{c|}{\textbf{Macro-F1}} &
        \multicolumn{1}{c}{\textbf{AUROC}} & 
        \multicolumn{1}{c}{\textbf{Accuracy}} &
        \multicolumn{1}{c}{\textbf{Macro-F1}} 
        \\ \hline
            ABMIL \cite{ilse2018abmil} & 92.5 (2.0) & 89.9 (2.8) & 89.8 (2.8) & 70.5 (23.9) & 69.0 (28.2) & 62.6 (23.4)  \\
            DSMIL \cite{li2021dsmil}  & 92.5 (1.7) & 87.3 (2.0) & 86.3 (2.0) & 81.5 (10.8) & 79.9 (18.5) & 72.7 (17.1) \\
            ReMix \cite{yang2022remix}  & 92.5 (7.2) & 90.0 (8.1) &  90.3 (7.7) & 75.0 (8.3) & 71.0 (14.3) & 67.9 (5.4) \\
            PatchGCN \cite{chen2021patchGCN}  & 88.6 (3.5)  & 92.1 (2.3) &  92.3 (2.4) & --- & --- & ---\\
            GTNMIL \cite{zheng2022GTNMIL} & 89.7 (4.7) & 81.2 (4.8) & 89.2 (4.9) & 61.8 (13.9) & 82.0 (2.8) & 62.3 (18.3)    \\
            H$^2$-MIL \cite{hou2022h2MIL}  & 92.1 (3.9)  & 88.2 (5.8)  & 88.0 (5.8) & 78.3 (11.1) & 87.0 (2.7) & 46.5 (0.8) \\
            HEAT \cite{chan2023HEAT}  & 92.8 (2.5) & 90.1 (4.5) & 90.3 (1.9) & 68.4  (6.5) & 86.1 (3.6) & 85.6 (13.3)\\
            BayesMIL \cite{yufei2022bayes} & 82.4 (9.5) & 78.3 (10.0)  & 80.3 (9.7) & 82.8 (5.2) & 85.1 (2.8) & 91.3 (1.6)  \\
             \hline
            \textbf{cDP-MIL} & \textbf{94.0 (1.7)} & \textbf{92.7 (1.1)} & \textbf{93.3 (0.7)} & \textbf{83.6 (0.2)} & \textbf{87.2 (0.2)} & \textbf{92.3 (0.1)}\\
        \bottomrule
        \end{tabular}}
    \label{tab: results_typing_ESCA}
\end{table}
\begin{table*}[h]
    \caption{Cancer classification performance of our method and other methods on the TCGA--COAD, TCGA--BRCA, and Camelyon 16 datasets, with standard deviations in brackets.}
    \centering
    \scalebox{0.65}{
        \begin{tabular}{lccc|ccc|ccc p{1.2cm}}
        \toprule
                \multicolumn{1}{l}{\textbf{}} &
        \multicolumn{3}{c|}{\textbf{Camelyon 16}} &
        \multicolumn{3}{c|}{\textbf{TCGA--COAD}} & 
        \multicolumn{3}{c}{\textbf{TCGA--BRCA}}  
        \\
        \multicolumn{1}{l}{\textbf{Model}} &
        \multicolumn{1}{c}{\textbf{AUROC}} & 
        \multicolumn{1}{c}{\textbf{Accuracy}} & 
        \multicolumn{1}{c|}{\textbf{F1}} &
        \multicolumn{1}{c}{\textbf{AUROC}} & 
        \multicolumn{1}{c}{\textbf{Accuracy}} & 
        \multicolumn{1}{c|}{\textbf{F1}} &
        \multicolumn{1}{c}{\textbf{AUROC}} & 
        \multicolumn{1}{c}{\textbf{Accuracy}} & 
        \multicolumn{1}{c}{\textbf{F1}} 
        \\ \hline
        ABMIL \cite{ilse2018abmil} & 95.9 (3.6) & 93.0 (6.7) & 91.1 (4.9) & 95.9 (2.3) & 95.9 (0.9) & 97.7 (2.2) & 97.7 (1.7) & 95.7 (1.1) & 97.4 (1.6) \\
        DSMIL \cite{li2021dsmil}  & 95.5 (4.3) & 94.0 (4.5) & 92.3 (4.9) & 93.4 (0.2) & 95.4 (0.5) & 97.5 (0.9) & 95.8 (0.5) & 96.3 (1.4) & 97.8 (2.0)\\
        Remix \cite{yang2022remix}  & 91.0 (6.0) & 92.5 (4.6) & 92.5 (4.7) & 94.3 (3.4) & 96.0 (4.6) & 92.8 (5.9) & 96.1 (0.7) & 95.8 (2.6) & 93.0 (3.4) \\
        PatchGCN \cite{chen2021patchGCN} & 69.9 (2.1) & 81.3 (3.2) & 88.9 (1.5) & 91.1 (5.3) & 97.1 (2.0) & 98.8 (1.0) & 96.2 (1.7) & 95.8 (2.6) & 93.0 (3.4)  \\ 
        GTNMIL \cite{zheng2022GTNMIL} & 57.9 (5.8) & 54.3 (3.6) & 51.8 (9.9) & 97.3 (2.6) & 98.1 (1.3) & 95.9 (2.4) & 94.7 (1.0) & 94.5 (0.2) & 93.7 (1.7) \\
        H$^2$MIL \cite{hou2022h2MIL} & 83.1 (11.1) & 81.9 (10.7) & 79.0 (13.4) & 99.7 (0.4) & 99.2 (0.5)  &  97.4 (1.7) & 97.9 (2.7) & 98.0 (1.5) & 97.6 (2.2)  \\ 
        HEAT \cite{chan2023HEAT} & 78.3 (0.4) & 84.4 (0.5) & 80.0 (0.3) & \textbf{99.7 (0.2)} & \textbf{99.4 (0.3)} & 98.7 (0.4) & 98.8 (0.7) & 98.3 (0.5) & 98.7 (0.7)  \\
        BayesMIL \cite{yufei2022bayes} & 94.6 (4.0) & 87.5 (4.1) & 82.2 (7.3) & 98.7 (1.4) & 96.9 (2.3) & 98.2 (1.3) & 99.5 (0.03) & 98.1 (0.4) & 98.3 (0.2)  \\
        \hline
        \textbf{cDP-MIL}  & \textbf{98.2 (0.01)} & \textbf{94.8 (0.6)} & \textbf{93.5 (0.6)} & \textbf{99.7 (0.2)} & \textbf{99.4 (0.1)} & \textbf{98.9 (0.1)} & \textbf{99.8 (0.02)}  & \textbf{98.5 (0.4)} & \textbf{99.2 (0.2)}\\
        \bottomrule
        \end{tabular}}
    \label{tab: cancer-classification}
    \vspace{-4mm}
\end{table*}

\begin{table}
\vspace{-4mm}
    \centering
         \caption{Performance [\%] of cDP-MIL on out-of-distribution (OOD) detection (compared to baseline uncertainty measures) and OOD generalization, with TCGA--COAD as the in-distribution dataset and TCGA--BRCA as the OOD dataset.
         }
         \scalebox{0.95}{
        \begin{tabular}{lcc||lcc p{1.3cm}}
        \toprule
        \multicolumn{1}{l}{\textbf{}} &
        \multicolumn{2}{c||}{\textbf{OOD Detection}} & \multicolumn{1}{l}{\textbf{}} &
        \multicolumn{3}{c}{\textbf{OOD Generalization}} 
        \\
        \multicolumn{1}{l}{\textbf{Measure}} &
        \multicolumn{1}{c}{\textbf{AUROC}} & 
        \multicolumn{1}{c||}{\textbf{AUPR}} &
        \multicolumn{1}{l}{\textbf{Model}} &
        \multicolumn{1}{l}{\textbf{AUROC}} &
        \multicolumn{1}{c}{\textbf{Accuracy}} &
        \multicolumn{1}{c}{\textbf{F1}}
        \\ \hline
            Max Confidence & 76.55 & 82.92 &  ABMIL \cite{ilse2018abmil} & 59.12 & 83.78 & 91.12 \\
            Entropy & 75.25 & 80.67 & DSMIL \cite{li2021dsmil} & 72.91 & 81.38 & 88.82 \\
            Differential Entropy & 68.39 & 86.54 & Remix \cite{yang2022remix} & 50.65 & 85.37 & 91.78\\
            \hline
            \textbf{Log Resp.} & \textbf{83.85} & \textbf{91.19} & \textbf{cDP-MIL} & \textbf{85.31} & \textbf{89.89} & \textbf{94.24}\\
            % \hline
        \bottomrule
        \end{tabular}
        }
    \label{tab: ood_detection}
    \vspace{-6mm}
\end{table}

\subsection{Out-of-Distribution Detection and Generalizability}
Moreover, we validate the ability of our model in uncertainty estimation (by detecting OOD samples) and generalizability (by generalization to OOD datasets), as shown in Table \ref{tab: ood_detection}.
We observe that using the log-likelihood from the posterior responsibility gives the best estimate of the uncertainty, and our method has satisfactory generalizability to other datasets compared with SOTA methods \cite{ilse2018abmil, li2021dsmil, yang2022remix}.

\subsection{Qualitative Results} 
\begin{figure}
\vspace{-5mm}
    \centering
\includegraphics[width=0.96\textwidth]{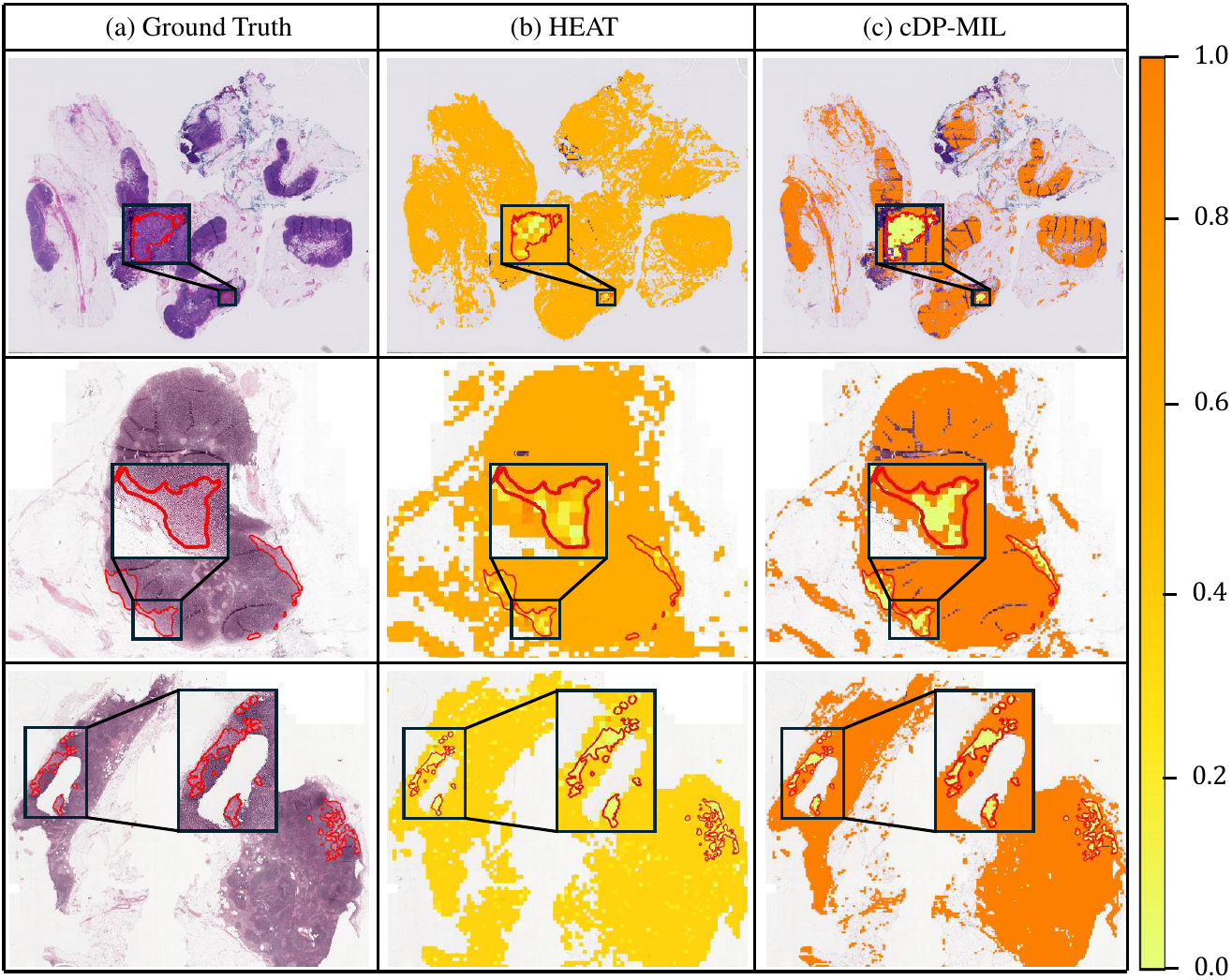}
    \caption{Tumor region localization.
    Left: Ground truth; middle: HEAT; right: cDP-MIL. 
    Ground truth regions are outlined with red boundaries.
    Lighter yellow indicates more important regions.
    }
    \label{fig: Localization}
    \vspace{-10mm}
\end{figure}

\vspace{0mm}
\label{subsect: qualtative}
We also study the interpretability performance with the localization task on the Camelyon 16 dataset.
In Fig. \ref{fig: Localization}, we provide a visualization of the patches in WSIs along with their corresponding scores. 
Our method can accurately locate tumor regions, as indicated by the ground truth annotations. 
It also achieves better localization performance compared to HEAT \cite{chan2023HEAT}, which
highlights its effectiveness in predicting instance-level labels.
\begin{figure}
    \centering
    \includegraphics[width=0.96\textwidth]{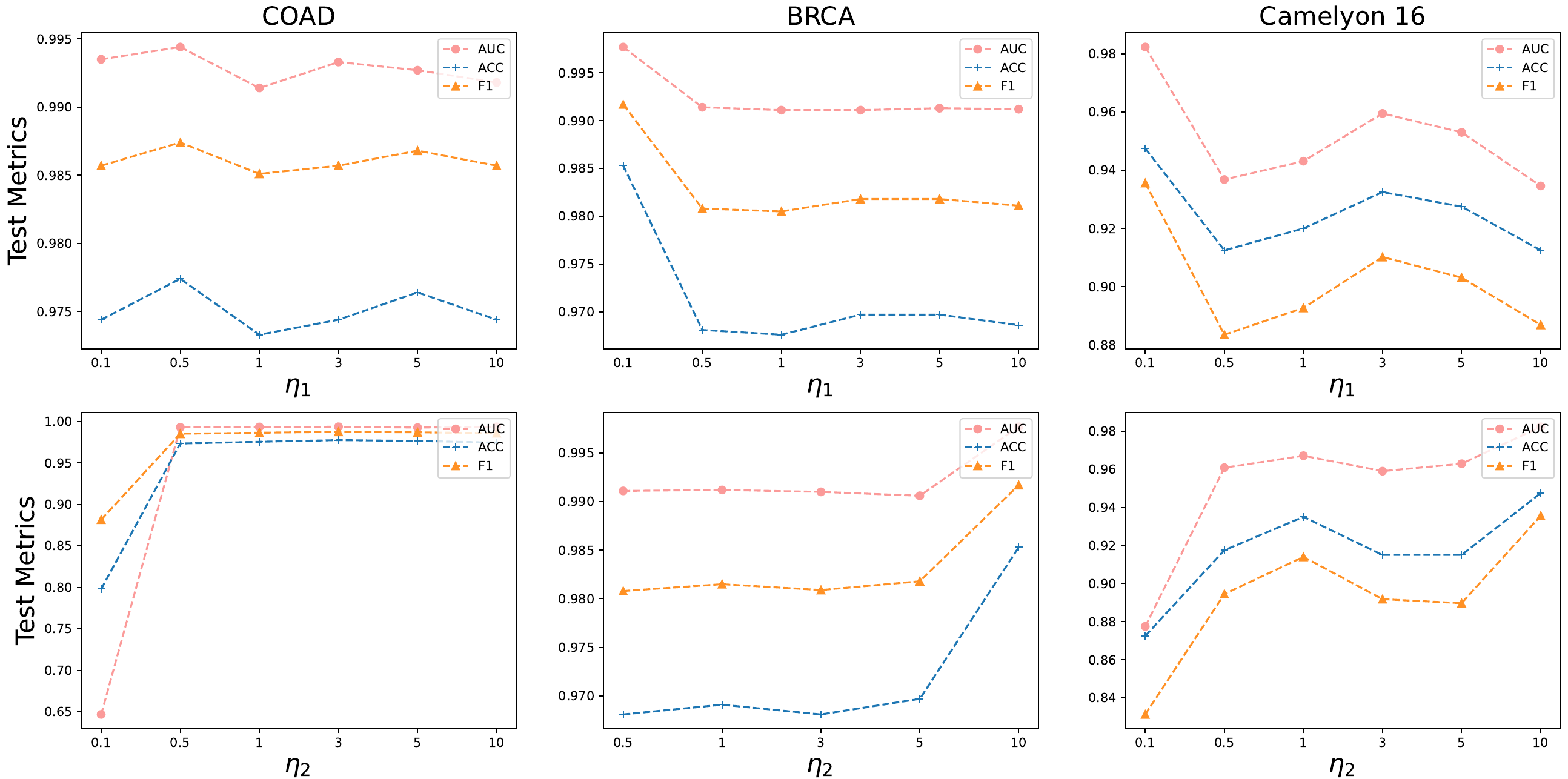}
    \caption{Performance of cDP-MIL with different concentration parameters $\eta_1$ and $\eta_2$. Left: COAD; middle: BRCA; right: Camelyon 16.}
    \label{fig: Ablation_Concentration}
\end{figure}

\begin{table}[t]
    \centering
         \caption{Performance [\%] of different pooling methods using the same DP prediction module on the Camelyon 16 dataset for cancer binary classification.}
        \begin{tabular}{lccc p{1.3cm}}
        \toprule
        \multicolumn{1}{l}{\textbf{Method}} &
        \multicolumn{1}{c}{\textbf{AUROC}} & 
        \multicolumn{1}{c}{\textbf{Accuracy}} &
        \multicolumn{1}{c}{\textbf{F1}} 
        \\ \hline
            Mean Pool & 77.6 (2.9) & 72.3 (1.4) & 66.1 (4.2)  \\
            Max Pool & 96.3 (0.7) & 90.8 (1.4) & 87.7 (1.6)  \\
            PL Pool & 78.3 (2.3) & 84.3 (4.3) & 80.0 (6.2)\\
            $k$-means  & 91.6 (1.5) & 87.8 (1.0) &  83.8 (1.9) \\ \hline
            \textbf{cDP-MIL} & \textbf{98.2 (0.02)} & \textbf{94.8 (0.6)} & \textbf{93.5 (0.6)} \\
        \bottomrule
        \end{tabular}
    \label{tab: Ablation_Pooling_Method}
    \vspace{-2mm}
\end{table}

\subsection{Ablation Analysis}
\para{Compared with Different Aggregation Algorithms.}
We validate the effectiveness of DP aggregation of cDP-MIL by comparing it with different aggregation methods including $k$-means, max pooling, mean pooling, and PL pool \cite{chan2023HEAT}.
We change only the aggregation methods while keeping other components intact. 
As shown in Table \ref{tab: Ablation_Pooling_Method}, our proposed DP pooling method obtains superior performance over the baseline pooling methods, which validates its contribution to the framework.

\begin{table*}[h]
    \vspace{-3mm}
    \caption{Performance [\%] of MLP and DP Classifier using the same DP aggregation module on the Camelyon 16, TCGA--COAD, and TCGA--BRCA datasets for cancer binary classification.}
    \centering
    \scalebox{0.75}{
        \begin{tabular}{lccc|ccc|ccc p{1.2cm}}
        \toprule
        \multicolumn{1}{l}{\textbf{}} &
        \multicolumn{3}{c}{\textbf{Camelyon 16}} &
        \multicolumn{3}{|c|}{\textbf{TCGA--COAD}} & 
        \multicolumn{3}{c}{\textbf{TCGA--BRCA}} 
        \\
        \multicolumn{1}{l}{\textbf{Classifier}} &
        \multicolumn{1}{c}{\textbf{AUROC}} & 
        \multicolumn{1}{c}{\textbf{Accuracy}} & 
        \multicolumn{1}{c|}{\textbf{F1}} &
        \multicolumn{1}{c}{\textbf{AUROC}} & 
        \multicolumn{1}{c}{\textbf{Accuracy}} & 
        \multicolumn{1}{c|}{\textbf{F1}} &
        \multicolumn{1}{c}{\textbf{AUROC}} & 
        \multicolumn{1}{c}{\textbf{Accuracy}} & 
        \multicolumn{1}{c}{\textbf{F1}} 
        \\ \hline
        MLP & 96.3 & 92.5 & 90.6 & 96.6 & 96.9 & 98.3 & 96.9 & 94.4 & 96.7 \\
        \hline
        DP Classifier & \textbf{98.2} & \textbf{94.8}  & \textbf{93.5} & \textbf{99.7} & \textbf{99.4} & \textbf{98.8} & \textbf{99.8} & \textbf{98.5} & \textbf{99.2} \\
        \bottomrule
        \end{tabular}}
    \label{tab: Ablation_Classifier}
    \vspace{-5mm}
\end{table*}

\para{Effect of DP Classifier.}
We also study the contribution of our proposed DP classifier when predicting the bag-level labels.
We compare our DP classifier with MLP, which is commonly used in MIL models. 
Table \ref{tab: Ablation_Classifier} shows that our proposed DP classifier (parameterized with the neural network) obtains a satisfactory performance, which validates the contribution of this proposed module. 
Note that our DP classifier can also accurately estimate the predictive uncertainty.

\para{Concentration parameters $\eta_1$ and $\eta_2$.}
The concentration parameters $\eta_1$ and $\eta_2$ are those of beta distributions in the stick-breaking process, which determines how actively we initiate new clusters when performing pooling and predictions. 
The smaller the $\eta$, the higher the probability that most points concentrate in only a few clusters.
Fig. \ref{fig: Ablation_Concentration} presents the change in performance as we take different values of $\eta$.
We observe that our method is in general robust to changes in $\eta$. 
For extremely small $\eta$, the model concentrates on the few top clusters quickly without actively exploring other potential clusters, leading to suboptimal performance.
When $\eta$ becomes larger, the instance embeddings would eventually converge to the top clusters (i.e., clusters with the highest likelihood), since the patches in WSIs mostly have distinctive characteristics (e.g., tumor vs normal patches).
Therefore, $\eta_1$ and $\eta_2$ are fine-grained hyperparameters which do not affect the learning performance much. 

\para{Effects of Maximum Number of Mixture Components $T$. }
We study the effects of the maximum number of components of the DP aggregation module. 
\begin{wrapfigure}{r}{0.48\textwidth}
    \centering
    \includegraphics[width=0.32\textwidth]{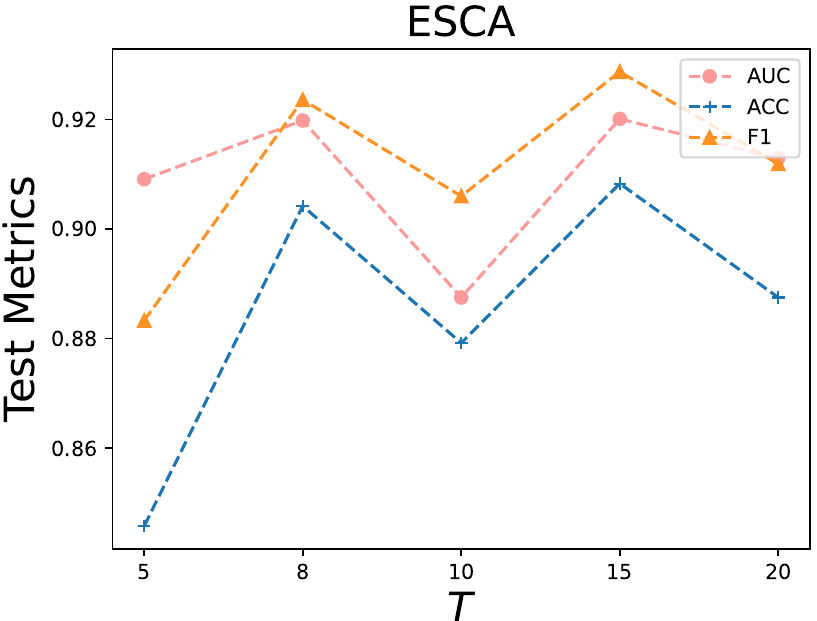}
    \caption{Performance of cDP-MIL on the ESCA dataset with different values of $T$.}
    \label{fig: Ablation_Component}
    \vspace{-4mm}
\end{wrapfigure}
Fig. \ref{fig: Ablation_Component} presents the results of our method on TCGA--ESCA with different numbers of maximum mixture components $T$. 
We observe the performance varies less when the maximum number of components is set high.
Unlike conventional clustering methods  (e.g., $k$-means) that require a predefined number of components, clustering with DP can automatically select the number of clusters.
This reduces the reliance on the number of components and improves the flexibility in feature pooling, which leads to more accurate bag-level prediction.

\section{Conclusion}

To solve the problem of multiple instance learning, we propose a novel framework based on a cascade of Dirichlet processes.
The DP aggregation module on the patch level can capture distributional characteristics of features more accurately by utilizing high-order information (e.g., covariance) while scaling up to large datasets through stochastic variational inference. 
In addition, our slide-level DP enables uncertainty estimation while maintaining satisfactory performance. 
The empirical results show that our method achieves superior performance to the state-of-the-art WSI analysis methods and accurately estimates uncertainty to detect OOD samples.
\section*{Acknowledgement}
This work was partially supported by the Research Grants Council of Hong Kong (27206123 and T45-401/22-N), the Hong Kong Innovation and Technology Fund (ITS/274/22), and the National Natural Science Foundation of China (No. 62201483).

\bibliographystyle{splncs04}
\bibliography{references}

\begin{thebibliography}{10}
\providecommand{\url}[1]{\texttt{#1}}
\providecommand{\urlprefix}{URL }
\providecommand{\doi}[1]{https://doi.org/#1}

\bibitem{andrews2002support}
Andrews, S., Tsochantaridis, I., Hofmann, T.: Support vector machines for multiple-instance learning. Advances in Neural Information Processing Systems  \textbf{15} (2002)

\bibitem{antoniak1974mixtures}
Antoniak, C.E.: Mixtures of dirichlet processes with applications to bayesian nonparametric problems. The Annals of Statistics pp. 1152--1174 (1974)

\bibitem{bejnordi2017camelyon16}
Bejnordi, B.E., Veta, M., Van~Diest, P.J., Van~Ginneken, B., Karssemeijer, N., Litjens, G., Van Der~Laak, J.A., Hermsen, M., Manson, Q.F., Balkenhol, M., et~al.: Diagnostic assessment of deep learning algorithms for detection of lymph node metastases in women with breast cancer. JAMA  \textbf{318}(22),  2199--2210 (2017)

\bibitem{bianchi2019mincut}
Bianchi, F.M., Grattarola, D., Alippi, C.: Mincut pooling in graph neural networks  (2019)

\bibitem{blei2006variationalDP}
Blei, D.M., Jordan, M.I.: Variational inference for dirichlet process mixtures. Bayesian Analysis  \textbf{1}(1),  121--144 (2006)

\bibitem{brancati2022bracs}
Brancati, N., Anniciello, A.M., Pati, P., Riccio, D., Scognamiglio, G., Jaume, G., De~Pietro, G., Di~Bonito, M., Foncubierta, A., Botti, G., et~al.: Bracs: A dataset for breast carcinoma subtyping in h\&e histology images. Database  \textbf{2022},  baac093 (2022)

\bibitem{chan2023HEAT}
Chan, T.H., Cendra, F.J., Ma, L., Yin, G., Yu, L.: Histopathology whole slide image analysis with heterogeneous graph representation learning. In: Proceedings of the IEEE/CVF Conference on Computer Vision and Pattern Recognition. pp. 15661--15670 (2023)

\bibitem{chan2024adaptive}
Chan, T.H., Lau, K.W., Shen, J., Yin, G., Yu, L.: Adaptive uncertainty estimation via high-dimensional testing on latent representations. Advances in Neural Information Processing Systems  \textbf{36} (2024)

\bibitem{chan2023SUMSHINE2}
Chan, T.H., Wong, C.H., Shen, J., Yin, G.: Source-aware embedding training on heterogeneous information networks. Data Intelligence  \textbf{5}(3),  611--635 (2023)

\bibitem{chen2022HIPT}
Chen, R.J., Chen, C., Li, Y., Chen, T.Y., Trister, A.D., Krishnan, R.G., Mahmood, F.: Scaling vision transformers to gigapixel images via hierarchical self-supervised learning. In: Proceedings of the IEEE/CVF Conference on Computer Vision and Pattern Recognition. pp. 16144--16155 (2022)

\bibitem{chen2021whole}
Chen, R.J., Lu, M.Y., Shaban, M., Chen, C., Chen, T.Y., Williamson, D.F., Mahmood, F.: Whole slide images are 2d point clouds: Context-aware survival prediction using patch-based graph convolutional networks. In: Medical Image Computing and Computer Assisted Intervention--MICCAI 2021: 24th International Conference, Strasbourg, France, September 27--October 1, 2021, Proceedings, Part VIII 24. pp. 339--349. Springer (2021)

\bibitem{chen2021patchGCN}
Chen, R.J., Lu, M.Y., Shaban, M., Chen, C., Chen, T.Y., Williamson, D.F., Mahmood, F.: Whole slide images are 2d point clouds: Context-aware survival prediction using patch-based graph convolutional networks. In: International Conference on Medical Image Computing and Computer-Assisted Intervention. pp. 339--349. Springer (2021)

\bibitem{chen2021multimodal}
Chen, R.J., Lu, M.Y., Weng, W.H., Chen, T.Y., Williamson, D.F., Manz, T., Shady, M., Mahmood, F.: Multimodal co-attention transformer for survival prediction in gigapixel whole slide images. In: Proceedings of the IEEE/CVF International Conference on Computer Vision. pp. 4015--4025 (2021)

\bibitem{chen2020simCLR}
Chen, T., Kornblith, S., Norouzi, M., Hinton, G.: A simple framework for contrastive learning of visual representations. In: International Conference on Machine Learning. pp. 1597--1607. PMLR (2020)

\bibitem{chen2023transformerMultimodal}
Chen, Y., Zhao, W., Yu, L.: Transformer-based multimodal fusion for survival prediction by integrating whole slide images, clinical, and genomic data. In: 2023 IEEE 20th International Symposium on Biomedical Imaging (ISBI). pp.~1--5. IEEE (2023)

\bibitem{chen2023TBMF}
Chen, Y., Zhao, W., Yu, L.: Transformer-based multimodal fusion for survival prediction by integrating whole slide images, clinical, and genomic data. In: 2023 IEEE 20th International Symposium on Biomedical Imaging (ISBI). pp.~1--5. IEEE (2023)

\bibitem{mmselfsup2021}
Contributors, M.: {MMSelfSup}: Openmmlab self-supervised learning toolbox and benchmark. \url{https://github.com/open-mmlab/mmselfsup} (2021)

\bibitem{dosovitskiy2021image}
Dosovitskiy, A., Beyer, L., Kolesnikov, A., Weissenborn, D., Zhai, X., Unterthiner, T., Dehghani, M., Minderer, M., Heigold, G., Gelly, S., Uszkoreit, J., Houlsby, N.: An image is worth 16x16 words: Transformers for image recognition at scale (2021)

\bibitem{echraibi2020DPGMM}
Echraibi, A., Flocon-Cholet, J., Gosselin, S., Vaton, S.: On the variational posterior of dirichlet process deep latent gaussian mixture models. In: ICML Workshop on Invertible Neural Networks, Normalizing Flows, and Explicit Likelihood Models (2020)

\bibitem{ferguson1973bayesian}
Ferguson, T.S.: A bayesian analysis of some nonparametric problems. The Annals of Statistics pp. 209--230 (1973)

\bibitem{gal2016dropout}
Gal, Y., Ghahramani, Z.: Dropout as a bayesian approximation: Representing model uncertainty in deep learning. In: International Conference on Machine Learning. pp. 1050--1059. PMLR (2016)

\bibitem{gao2019graph}
Gao, H., Ji, S.: Graph u-nets. In: International Conference on Machine Learning. pp. 2083--2092. PMLR (2019)

\bibitem{graham2019hover}
Graham, S., Vu, Q.D., Raza, S.E.A., Azam, A., Tsang, Y.W., Kwak, J.T., Rajpoot, N.: Hover-net: Simultaneous segmentation and classification of nuclei in multi-tissue histology images. Medical Image Analysis  \textbf{58},  101563 (2019)

\bibitem{Gustafsson_2020_CVPR_Workshops}
Gustafsson, F.K., Danelljan, M., Schon, T.B.: Evaluating scalable bayesian deep learning methods for robust computer vision. In: Proceedings of the IEEE/CVF Conference on Computer Vision and Pattern Recognition (CVPR) Workshops (June 2020)

\bibitem{haussmann2017VGPMIL}
Hau{\ss}mann, M., Hamprecht, F.A., Kandemir, M.: Variational bayesian multiple instance learning with gaussian processes. In: Proceedings of the IEEE Conference on Computer Vision and Pattern Recognition. pp. 6570--6579 (2017)

\bibitem{he2020Moco}
He, K., Fan, H., Wu, Y., Xie, S., Girshick, R.: Momentum contrast for unsupervised visual representation learning. In: Proceedings of the IEEE/CVF conference on computer vision and pattern recognition. pp. 9729--9738 (2020)

\bibitem{hoffman2013SVI}
Hoffman, M.D., Blei, D.M., Wang, C., Paisley, J.: Stochastic variational inference. Journal of Machine Learning Research  (2013)

\bibitem{hou2022h2MIL}
Hou, W., Yu, L., Lin, C., Huang, H., Yu, R., Qin, J., Wang, L.: H2-mil: Exploring hierarchical representation with heterogeneous multiple instance learning for whole slide image analysis. In: Thirty-sixth AAAI conference on artificial intelligence (2022)

\bibitem{hughes2013memoized}
Hughes, M.C., Sudderth, E.: Memoized online variational inference for dirichlet process mixture models. Advances in Neural Information Processing Systems  \textbf{26} (2013)

\bibitem{ilse2018abmil}
Ilse, M., Tomczak, J., Welling, M.: Attention-based deep multiple instance learning. In: International Conference on Machine Learning. pp. 2127--2136. PMLR (2018)

\bibitem{ishwaran2002exactDP}
Ishwaran, H., Zarepour, M.: Exact and approximate sum representations for the dirichlet process. Canadian Journal of Statistics  \textbf{30}(2),  269--283 (2002)

\bibitem{jiang2016variational}
Jiang, Z., Zheng, Y., Tan, H., Tang, B., Zhou, H.: Variational deep embedding: A generative approach to clustering. Proceedings of the Twenty-Sixth International Joint Conference on Artificial Intelligence  (2017)

\bibitem{kandasamy2018neural}
Kandasamy, K., Neiswanger, W., Schneider, J., Poczos, B., Xing, E.P.: Neural architecture search with bayesian optimisation and optimal transport. Advances in Neural Information Processing Systems  \textbf{31} (2018)

\bibitem{kandemir2014DPMIL}
Kandemir, M., Hamprecht, F.A., et~al.: Instance label prediction by dirichlet process multiple instance learning. In: Proceddings of The Conference on Uncertainty in Artificial Intelligence. pp. 380--389 (2014)

\bibitem{kendall2017uncertainties}
Kendall, A., Gal, Y.: What uncertainties do we need in bayesian deep learning for computer vision? Advances in Neural Information Processing Systems  \textbf{30} (2017)

\bibitem{kim2010gaussian}
Kim, M., De~la Torre, F.: Gaussian processes multiple instance learning. In: International Conference on Machine Learning. pp. 535--542. Citeseer (2010)

\bibitem{kingma2013auto}
Kingma, D.P., Welling, M.: Auto-encoding variational bayes. arXiv preprint arXiv:1312.6114  (2013)

\bibitem{li2021dsmil}
Li, B., Li, Y., Eliceiri, K.W.: Dual-stream multiple instance learning network for whole slide image classification with self-supervised contrastive learning. In: Proceedings of the IEEE/CVF Conference on Computer Vision and Pattern Recognition. pp. 14318--14328 (2021)

\bibitem{li2020gp}
Li, Z., Xi, T., Deng, J., Zhang, G., Wen, S., He, R.: Gp-nas: Gaussian process based neural architecture search. In: Proceedings of the IEEE/CVF Conference on Computer Vision and Pattern Recognition. pp. 11933--11942 (2020)

\bibitem{liu2023dsca}
Liu, P., Fu, B., Ye, F., Yang, R., Ji, L.: Dsca: A dual-stream network with cross-attention on whole-slide image pyramids for cancer prognosis. Expert Systems with Applications  \textbf{227},  120280 (2023)

\bibitem{lu2021CLAM}
Lu, M.Y., Williamson, D.F., Chen, T.Y., Chen, R.J., Barbieri, M., Mahmood, F.: Data-efficient and weakly supervised computational pathology on whole-slide images. Nature Biomedical Engineering  \textbf{5}(6),  555--570 (2021)

\bibitem{malinin2018DPN}
Malinin, A., Gales, M.: Predictive uncertainty estimation via prior networks. Advances in Neural Information Processing Systems  \textbf{31} (2018)

\bibitem{mauldin1992polya}
Mauldin, R.D., Sudderth, W.D., Williams, S.C.: Polya trees and random distributions. The Annals of Statistics pp. 1203--1221 (1992)

\bibitem{rasmussen2006gaussian}
Rasmussen, C.E., Williams, C.K., et~al.: Gaussian processes for machine learning, vol.~1. Springer (2006)

\bibitem{rezende2014stochastic}
Rezende, D.J., Mohamed, S., Wierstra, D.: Stochastic backpropagation and approximate inference in deep generative models. In: International Conference on Machine Learning. pp. 1278--1286. PMLR (2014)

\bibitem{ronen2022deepdpm}
Ronen, M., Finder, S.E., Freifeld, O.: Deepdpm: Deep clustering with an unknown number of clusters. In: Proceedings of the IEEE/CVF Conference on Computer Vision and Pattern Recognition. pp. 9861--9870 (2022)

\bibitem{shah2018deep}
Shah, S.A., Koltun, V.: Deep continuous clustering. arXiv preprint arXiv:1803.01449  (2018)

\bibitem{shao2021transmil}
Shao, Z., Bian, H., Chen, Y., Wang, Y., Zhang, J., Ji, X., et~al.: Transmil: Transformer based correlated multiple instance learning for whole slide image classification. Advances in Neural Information Processing Systems  \textbf{34},  2136--2147 (2021)

\bibitem{snoek2012practical}
Snoek, J., Larochelle, H., Adams, R.P.: Practical bayesian optimization of machine learning algorithms. Advances in Neural Information Processing Systems  \textbf{25} (2012)

\bibitem{Wang_2022_CVPR}
Wang, J., Lukasiewicz, T.: Rethinking bayesian deep learning methods for semi-supervised volumetric medical image segmentation. In: Proceedings of the IEEE/CVF Conference on Computer Vision and Pattern Recognition (CVPR). pp. 182--190 (June 2022)

\bibitem{wang2019rmdl}
Wang, S., Zhu, Y., Yu, L., Chen, H., Lin, H., Wan, X., Fan, X., Heng, P.A.: Rmdl: Recalibrated multi-instance deep learning for whole slide gastric image classification. Medical Image Analysis  \textbf{58},  101549 (2019)

\bibitem{wang2021dnb}
Wang, Z., Ni, Y., Jing, B., Wang, D., Zhang, H., Xing, E.: Dnb: A joint learning framework for deep bayesian nonparametric clustering. IEEE Transactions on Neural Networks and Learning Systems  \textbf{33}(12),  7610--7620 (2021)

\bibitem{weinstein2013TCGA}
Weinstein, J.N., Collisson, E.A., Mills, G.B., Shaw, K.R., Ozenberger, B.A., Ellrott, K., Shmulevich, I., Sander, C., Stuart, J.M.: The cancer genome atlas pan-cancer analysis project. Nature genetics  \textbf{45}(10),  1113--1120 (2013)

\bibitem{yang2017towards}
Yang, B., Fu, X., Sidiropoulos, N.D., Hong, M.: Towards k-means-friendly spaces: Simultaneous deep learning and clustering. In: International Conference on Machine Learning. pp. 3861--3870. PMLR (2017)

\bibitem{yang2022remix}
Yang, J., Chen, H., Zhao, Y., Yang, F., Zhang, Y., He, L., Yao, J.: Remix: A general and efficient framework for multiple instance learning based whole slide image classification. Medical Image Computing and Computer Assisted Intervention--MICCAI 2022: 25th International Conference, Singapore, September 18--22, 2022, Proceedings, Part II  (2022)

\bibitem{yang2019deep}
Yang, L., Cheung, N.M., Li, J., Fang, J.: Deep clustering by gaussian mixture variational autoencoders with graph embedding. In: Proceedings of the IEEE/CVF international conference on computer vision. pp. 6440--6449 (2019)

\bibitem{Teh2006HDP}
Yee Whye~Teh, Michael I~Jordan, M.J.B., Blei, D.M.: Hierarchical dirichlet processes. Journal of the American Statistical Association  \textbf{101}(476),  1566--1581 (2006). \doi{10.1198/016214506000000302}, \url{https://doi.org/10.1198/016214506000000302}

\bibitem{yufei2022bayes}
Yufei, C., Liu, Z., Liu, X., Liu, X., Wang, C., Kuo, T.W., Xue, C.J., Chan, A.B.: Bayes-mil: A new probabilistic perspective on attention-based multiple instance learning for whole slide images. In: The Eleventh International Conference on Learning Representations (2022)

\bibitem{ZhangChenyang2023Bnao}
Zhang, C., Yin, G.: Bayesian nonparametric analysis of restricted mean survival time. Biometrics  \textbf{79}(2),  1383--1396 (2023)

\bibitem{zhao2023mulgt}
Zhao, W., Wang, S., Yeung, M., Niu, T., Yu, L.: Mulgt: Multi-task graph-transformer with task-aware knowledge injection and domain knowledge-driven pooling for whole slide image analysis. In: Thirty-seventh AAAI conference on artificial intelligence (2023)

\bibitem{zheng2022GTNMIL}
Zheng, Y., Gindra, R.H., Green, E.J., Burks, E.J., Betke, M., Beane, J.E., Kolachalama, V.B.: A graph-transformer for whole slide image classification. IEEE Transactions on Medical Imaging  \textbf{41}(11),  3003--3015 (2022)

\end{thebibliography}
\end{document}